\documentclass{article}

\usepackage[dblblindworkshop, final]{neurips_2026}
\workshoptitle{AI-Native Academia}

\usepackage{graphicx}
\usepackage[most]{tcolorbox}
\usepackage{xcolor}
\usepackage{tabularx}
\usepackage{booktabs}
\usepackage{xurl}
\usepackage{graphicx}
\usepackage[utf8]{inputenc} % allow utf-8 input
\usepackage[T1]{fontenc}    % use 8-bit T1 fonts
\usepackage{hyperref}       % hyperlinks
\usepackage{url}            % simple URL typesetting
\usepackage{amsfonts}       % blackboard math symbols
\usepackage{nicefrac}       % compact symbols for 1/2, etc.
\usepackage{microtype}      % microtypography

\title{Policy-Conditioned AI-Use Detection:\\
An Evidentiary Framework for Academic Publishing}

\author{%
Jairo~Diaz-Rodriguez
    \\
  Department of Mathematics and Statistics\\
  York University\\
  Toronto, Ontario M3J 1P3 \\
  \texttt{jdiazrod@yorku.ca}
\And
Mumin Jia
    \\
  Department of Mathematics and Statistics\\
  York University\\
  Toronto, Ontario M3J 1P3 \\
  \texttt{amyjia@yorku.ca}
}

\begin{document}

\maketitle

\begin{abstract}
Major venues now publish detailed rules about how authors, reviewers, and area chairs may use AI, and those rules differ by role, by task, and by what must be disclosed. AI detection, the instrument usually proposed to enforce them, estimates something else: whether an AI model wrote the text. We argue that this target is misaligned with the decisions conferences and journals face, and propose \emph{policy-conditioned AI-use detection}, an evidentiary framework for assessing whether a human--AI workflow complied with a stated rule. \textbf{Policy} makes the governing rule an explicit input. \textbf{Inference} reports hypotheses, evidence, calibration regime, and uncertainty in place of verdicts such as ``AI detected''. \textbf{Evaluation} builds benchmarks from reproducible pipelines that generate compliant and non-compliant workflows, and reports true positive rate at a false positive rate the venue fixes in advance. We work the framework through peer review, where at plausible violation rates a detector at a strong operating point still flags more compliant authors than violating ones. The framework therefore also names what a venue must instrument: structured disclosure, approved-tool routing that respects reviewer confidentiality, and a path by which a finding can be contested. Under this framing a detector is not an authorship classifier but an auditable procedure with an error rate the venue fixes in advance and can defend.
\end{abstract}

\section{Introduction}\label{sec:intro}

Academic venues have written detailed rules about how AI may be used, and have almost no way to tell whether those rules were followed. NeurIPS 2026 exempts spell checking, grammar suggestions, editing aid, and basic code assistance from any documentation requirement, and asks that agents or LLMs be described in the experimental setup when they form an important, original, or non-standard component of the approach. Its reviewers may use no LLM beyond a venue-sanctioned one, and only on papers where such use is specifically permitted \citep{neurips2026handbook}. ICML 2026 allows authors to use generative AI to assist in writing or research, and splits reviewer policy into two regimes: one prohibits LLM use outright, the other permits privacy-compliant help with comprehension and polish while forbidding any delegation of judgment \citep{icml2026cfp,icml2026review}. ICLR 2026 sets a lower threshold still and requires that any use of an LLM be declared, in the paper and on the submission form, with authors and reviewers remaining responsible for whatever a model contributes \citep{iclr2026author,iclr2026llm}. The three venues do not agree even on when disclosure is owed: routine language editing needs no mention at NeurIPS and must be declared at ICLR. These are not variations on a single prohibition. They are role-conditioned rules about delegation, disclosure, confidentiality, and retained responsibility.

\begin{figure}[t]
\centering
\includegraphics[width = 0.95\textwidth]{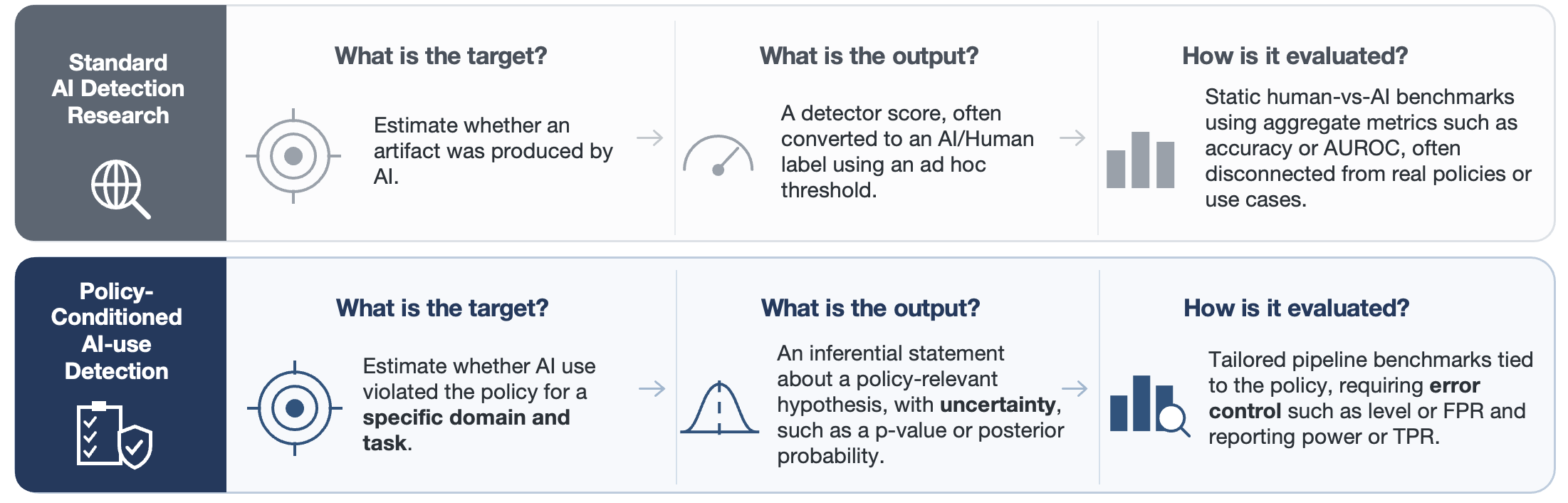}
\caption{\small{
Standard AI detection research versus policy-conditioned AI-use detection: what the detection target should be, what the output should represent, and how the system should be evaluated.}}
\label{fig:difference}
\vspace{-0.4cm}
\end{figure}

What venues have instead is AI detection research organized around origin classification: given an artifact, estimate whether a language model produced it \citep{gehrmann2019gltr,mitchell2023detectgpt,guo2023hc3,verma2024ghostbuster,wu2025survey}. That target made sense when model-written text was rare and separable from ordinary scholarly writing. It fits badly a pipeline in which an author may legitimately polish prose with a model, a reviewer may fix grammar but not form an opinion, and an area chair may cluster reviews but not set a decision. The mismatch is not a matter of accuracy. A reviewer who drafts a full assessment and asks a model to correct the English satisfies most venue rules; a reviewer who pastes a confidential manuscript into a consumer chatbot breaks them even if every word of the review is their own. An origin detector rates the first as the more suspicious of the two, inverting the rule it is meant to enforce. Large-scale estimates already show model-modified signal in a substantial share of review text at AI venues \citep{liang2024monitoring} and score effects associated with AI-assisted reviews \citep{latona2024lottery}, and neither finding tells a chair which of those reviews broke a rule. Under most current policies, some of them broke none.

The gap matters because the decisions on the other side of a detector score are consequential and contested. A flag can mean a desk rejection, an integrity investigation, a retracted review, or a note in an author's record. The accused has nothing to argue with: a number states no proposition, names no hypothesis, and identifies no population on which its threshold was calibrated. Scale compounds the problem: violations are rare and submissions many, so a strong operating point still accuses more compliant authors than violating ones, and no ranking metric reveals it. Detectors also fail in ways their users cannot anticipate, degrading under paraphrase and domain shift and penalizing non-native English writers \citep{sadasivan2023,weberwulff2023,liang2023}; OpenAI withdrew its own text classifier for low accuracy \citep{openai2023classifier}. Meanwhile the artifacts themselves are being written to defeat inspection, most visibly through hidden prompts embedded in manuscripts to steer AI-assisted reviewers \citep{nikkei2025hiddenprompt}.

\textbf{AI-detection research for academic publishing should change its target. The question is not whether an artifact contains AI, but whether the evidence available about the workflow that produced it is more consistent with a use the governing policy permits or with one it prohibits, for the specific role and task at hand (Figure~\ref{fig:difference}).} Origin detection does not disappear under this view. It becomes the special case in which the policy happens to be ``no AI-generated content.''

\begin{figure}[t]
\centering
\includegraphics[width = 0.90\textwidth]{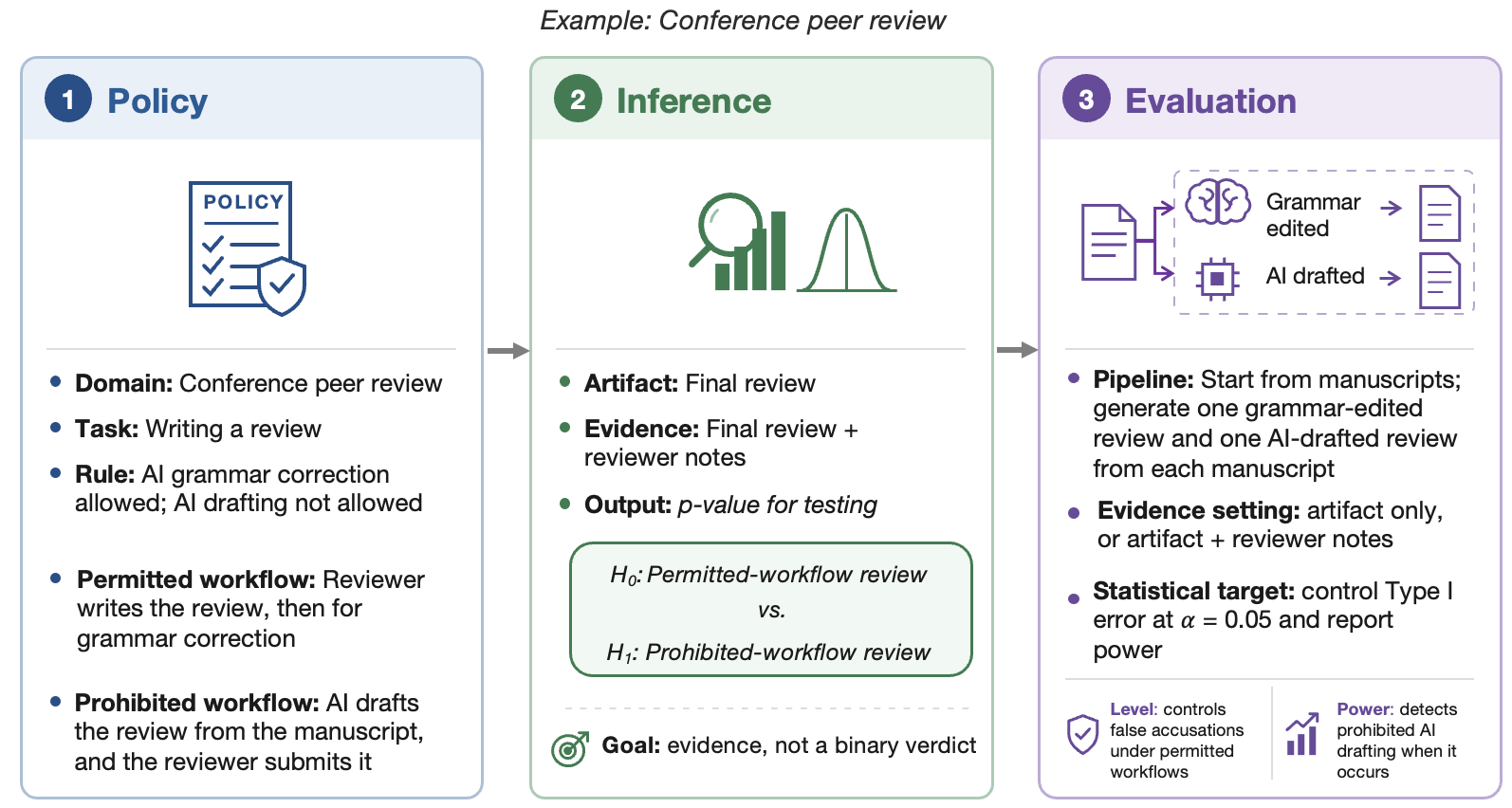}
\caption{\small{
Policy-conditioned AI-use detection connects three components, shown here for peer review at a venue that permits AI grammar correction but prohibits AI drafting of the review: \textbf{policy} defines the permitted and prohibited reviewer workflows, \textbf{inference} quantifies evidence for a policy violation, and \textbf{evaluation} measures TPR at controlled FPR levels.
}}
\label{fig:framework}
\vspace{-0.4cm}
\end{figure}

This shift motivates the three-part framework developed in the paper: {policy}, {inference}, and {evaluation}. First, \textbf{policy} changes the target of detection from AI use in general to policy-relevant AI use. The operative distinction is not human versus AI but permitted assistance versus prohibited delegation, and what counts as permitted depends on the domain, role, task, and governing rule. Second, \textbf{inference} changes the statistical form of detection. Outputs such as ``AI detected'' collapse uncertain evidence into an overconfident verdict, where a detector should instead report the hypotheses, evidence, uncertainty, calibration assumptions, and validity conditions behind its claim. Third, \textbf{evaluation} changes how detectors are assessed. Existing benchmarks score artifact labels rather than policy-relative workflow compliance, so the field needs pipeline benchmarks that start from human seeds and generate artifacts through controlled permitted and prohibited workflows, reporting power against those workflows at an error rate the venue has fixed in advance (Figure~\ref{fig:framework}).

The rest of the paper is organized as follows. Section~\ref{sec:framework} introduces the framework. Sections~\ref{sec:conceptual}, \ref{sec:stat_inf}, and \ref{sec:benchmark} develop each component in turn. Section~\ref{sec:governance} sets out what a venue would have to build for the framework to run. Section~\ref{sec:related_work} situates the framework within related work, and Section~\ref{sec:discussion} considers objections and limitations.

\section{A Framework for Policy-Conditioned AI-Use Detection}\label{sec:framework}

The framework has three components. \emph{Policy} fixes which forms of AI use are permitted for a given domain, role, and task. \emph{Inference} weighs the evidence about an observed artifact against those categories. \emph{Evaluation} tests whether the resulting procedure controls the errors that matter. None works alone: a detector without policy does not know what it is estimating, scores without inference cannot support accountable decisions, and benchmark numbers without appropriate evaluation do not transfer to deployment. The aim is not a universal classifier for ``AI'' versus ``human,'' but an evidentiary system for reasoning about whether a workflow complied with a stated rule under stated uncertainty.

\paragraph{Policy.}
Detection should begin with an explicit rule: a detector should receive not only an artifact but the domain, role, task, and policy that define permitted and prohibited use. Take a venue that allows reviewers to edit language but forbids delegating judgment. A permitted workflow has the reviewer form an assessment, write it, and ask a model to fix grammar; a prohibited one has the reviewer upload the manuscript and ask a model for the review and the recommendation. Appropriate use is not a property of the final review but a relation among a workflow, a role, and a policy.

\paragraph{Inference.}
Given a policy, the detector should produce an inferential statement rather than a label. The observed artifact is the submitted review; further evidence may include reviewer notes, portal edit history, tool logs, or a disclosure. The relevant output is evidence about
$$
\begin{aligned}
H_0 &: \text{the review was produced through a workflow permitted by the venue's policy},\\
H_1 &: \text{the review was produced through a workflow prohibited by that policy}.
\end{aligned}
$$
A frequentist report states the null, the test statistic, the calibration regime, and the p-value. A Bayesian report states the prior, the evidence used, and the posterior probability of a violation. In both cases uncertainty is explicit, and the detector's role is to supply evidence for human adjudication rather than to convert a score into an accusation.

\paragraph{Evaluation.}
Benchmarks should evaluate the full procedure, not artifact classification alone. Evaluation should start from a shared pool of manuscripts, generate several review workflows from each, and test the detector under different evidence conditions, such as artifact-only access versus artifact plus notes and logs. The statistical target must be stated: a benchmark may require Type~I error control at $\alpha = 0.05$ under permitted workflows while reporting TPR at that FPR against prohibited delegation. Reporting AUROC alone names neither the tolerated false-accusation rate nor the detection rate at that tolerance.

Together these define a research agenda: policy languages that specify permitted and prohibited AI use per role, detectors that combine artifact signals with provenance and workflow evidence, and benchmarks built around reproducible human--AI pipelines rather than frozen datasets. Appendix~\ref{app:review} instantiates the framework end to end for peer review, and Appendix~\ref{app:policyspec} sketches the corresponding machine-readable policy; the following sections develop each component.

\section{Policy: AI Use Is Not the Right Target}\label{sec:conceptual}

\subsection{Venue policy is already role- and task-specific}

The policies quoted in Section~\ref{sec:intro} share a structure, and it is not an administrative detail: it changes the quantity a detection system should be estimating. Authors may use AI and stay responsible for the result, reviewers face tighter constraints because review involves confidential material and delegable judgment, and disclosure attaches to significant or non-standard use rather than to any use at all. Venues have also begun deploying review-side AI themselves \citep{iclr2025feedback} and running experiments with AI authors and reviewers under explicit disclosure \citep{agents4science2025}, which makes the line between sanctioned and unsanctioned automation a matter of provenance rather than style.

The structure is not a quirk of one community: education, platform, and regulatory regimes condition obligations on task, data, tool, and form of generation rather than on AI use as such \citep{ib2023,harvard2026,euaiact2024}. Across all of them the operative distinction is allowed versus disallowed use, not use versus non-use.

\subsection{Appropriate use is a relation, not a property}

Because rules differ across domains and roles, identical AI participation can be required, permitted, irrelevant, or prohibited depending on context. Grammar editing is unremarkable in a manuscript and in a review, while undisclosed full drafting violates authorship norms in the first and delegates judgment in the second. In software engineering, agentic systems may generate entire pull requests when they operate through approved tools, tests, code review, and branch protections \citep{github2025agent,openai2026codex,google2025gemini}, and the same generation volume would be disqualifying in a written examination. Appropriate use is therefore not an intrinsic property of a text, figure, or code file. It is a relation between the artifact, the workflow that produced it, and the policy governing that workflow, and a detector that ignores two of the three terms is estimating something no institution asked about.

\subsection{Formalization}

Let $x$ denote the observed artifact, $d$ a domain, $t$ a task, $\rho$ the actor's role, $\pi_{d,t,\rho}$ the applicable policy, and $e$ the available evidence beyond the artifact. Lowercase symbols denote observed inputs to the detector. A detector should not estimate only whether $x$ is AI-generated; it should report evidence about whether the workflow that produced $x$ complies with $\pi_{d,t,\rho}$:
$$
D(x,d,t,\rho,\pi_{d,t,\rho},e) \rightarrow r,
$$
where $r$ is a structured report rather than a binary verdict. A report may contain a compliance assessment, a posterior probability of violation, uncertainty, the evidence types used and missing, the calibration regime, validity conditions, and a recommended action including abstention.

This makes three changes. The policy becomes an input to the detector instead of an afterthought. The estimand shifts from an artifact label to a workflow class. And the output is an evidentiary report, not an accusation.

How much inference this requires depends on how directly the workflow is observed. Under artifact-only access the production workflow is largely latent and compliance has to be inferred from indirect signals. Provenance metadata, disclosures, edit histories, and approved-tool logs each narrow the set of workflows consistent with what the venue holds. At the limit, a sufficiently complete trace establishes a policy-relevant action outright, and the task becomes verification instead of statistical reconstruction. We use \emph{policy-conditioned AI-use detection} as an umbrella term for the whole of this evidentiary procedure. A statistical detector is one evidence source within it, and in some evidence regimes not the dominant one.

\subsection{Policies must become machine-readable}

Conditioning on policy requires policy that a system can read. Venue rules are currently prose scattered across handbooks, calls for papers, and blog posts, which makes them impossible to attach to a submission, version, or audit. A minimal specification would name, for each role: the tasks in scope; the workflows classified as permitted and prohibited; the disclosure required for each; the evidence the venue will collect and retain; the error rate the venue is willing to tolerate against compliant actors; and the graduated consequences of a sustained finding. Appendix~\ref{app:policyspec} sketches such a specification for reviewer conduct. The benefit does not depend on any detector working well: a venue that writes the specification must state its tolerated false-accusation rate before it sees any cases, which is the only point at which that number can be chosen honestly.

\section{Inference: Detection as a Statistical Procedure}\label{sec:stat_inf}

\subsection{From detector scores to statistical claims}

The phrase ``AI detected'' is rhetorically strong and statistically empty. It suggests that a hidden property of the artifact was observed. In practice a detector emits a score, and a threshold turns that score into a label. The threshold is chosen by a vendor, an institution, or an individual user, usually with no stated relationship to the policy being enforced or to the error rate that matters in the setting.

Detectors are compared using AUROC, accuracy, and occasionally TPR at one operating point. None of these tells a program chair what score should trigger an integrity inquiry. \citet{tufts2025practical} identify the gap and argue for operating-point reporting. In a hypothesis-testing framing the natural summary is TPR at a fixed FPR, where FPR is the Type~I error rate under permitted workflows, because it names both the tolerated false-accusation rate and the detection rate obtained at that tolerance.

Deployment forces the decision either way. A venue must decide which score triggers review, which triggers nothing, and which is too uncertain to act on; absent a statistical framework those thresholds are set by intuition, and scores end up treated as evidence of misconduct with no error control. The defect is in the form of the output rather than in the model behind it. ``The detector returned 0.87'' asserts no proposition, so an author has nothing to rebut and a chair has nothing to explain. A more accurate detector reporting the same number leaves the venue where it started.

\subsection{Hypothesis testing}

One formalization is hypothesis testing. Even conventional origin detection is better stated as a test than as a verdict:
$$H_0: \text{the artifact } x \text{ is human-authored}
\quad \text{against} \quad
H_1: \text{the artifact } x \text{ is AI-generated}.$$
We argue that this should be generalized to policy-conditioned hypotheses:
\[
\begin{array}{ll}
H_0^{(d,t,\rho)}: & \text{the artifact } x \text{ was produced through a workflow permitted by } \pi_{d,t,\rho},\\
H_1^{(d,t,\rho)}: & \text{the artifact } x \text{ was produced through a workflow prohibited by } \pi_{d,t,\rho}.
\end{array}
\]
A thresholded score is not enough to support this. A valid test specifies the null, the alternatives, the statistic, the calibration distribution, the population over which calibration is expected to hold, and the level. Here Type~I error is the probability of rejecting a permitted workflow, so controlling the level bounds the probability that a compliant author or reviewer is flagged, and the required stringency depends on the consequence. If a flag only prioritizes low-stakes human reading, a weakly calibrated screen may be acceptable; if it can trigger a desk rejection, an integrity case, or removal from a reviewer pool, the venue should be able to state the level and the assumptions under which it holds. Failing to reject is not proof of compliance, and rejecting is not proof of misconduct; both are evidence under a stated model, calibration regime, and level.

\subsection{Bayesian inference}

A Bayesian framing is often more natural for deployment. Let $V$ denote the event that the production workflow violates the applicable policy. A detector may report
$$P(V \mid x, d, t, \rho, \pi_{d,t,\rho}, e).$$
The posterior depends on likelihoods and on priors, and the same score should imply different concern in different settings. Evidence of substantial model drafting is decisive in a closed-book assessment, weak evidence of anything in a marketing document, and expected in an approved agentic coding workflow. A watermark is uninformative if the policy allowed disclosed AI drafting and highly informative if it prohibited model-written prose.

Bayesian reporting also makes base rates and costs explicit. A detector score cannot be interpreted without an assumption about how common the relevant violation is. Institutions differ in their loss functions---a university may weight false accusations most heavily, a journal the fabricated citations that survive into the literature---and no such weighting is recoverable from a generic origin score. Abstention deserves the same first-class treatment: where violations are rare, a procedure that returns ``insufficient evidence'' on short artifacts with no workflow trace beats one that guesses, because the guesses are mostly wrong and each costs a compliant person something.

\paragraph{What a report should contain.}
A frequentist report states the null, the statistic, the calibration regime, and the p-value before recommending review or abstention. A Bayesian report states the prior assumptions, the evidence used and missing, and the posterior probability of violation. Instead of ``AI detected,'' a usable report reads: \emph{estimated probability of prohibited drafting is $0.38$; calibration is weak for reviews under 300 words}, or \emph{$p = 0.049$ in a level-$0.05$ test calibrated on long-form reviews from this venue's 2025 cycle}. These pair a number with its scope of validity, which makes them auditable and much harder to misread as proof. The goal is not to remove decision rules but to make them explicit, policy-specific, and contestable.

\section{Evaluation: Benchmarks Should Be Recipes, Not Datasets}\label{sec:benchmark}

\subsection{Current benchmarks and their limits}

Machine-generated text detection has been evaluated largely on static corpora that separate human writing from model output or attribute text to a source generator \citep{uchendu2021turingbench,guo2023hc3,he2023mgtbench,wang2024m4,li2024mage}. A second line starts from human text and applies model transformations, better reflecting workflows in which people edit, paraphrase, summarize, expand, or mix human and model prose, and studies adversarial rewriting, mixed authorship, boundary detection, and graded editing \citep{su2023hc3plus,hu2023radar,koike2024outfox,dugan2024raid,wang2024semeval,thai2026editlens}.

These are real advances, and they remain insufficient for governance. Benchmark tasks are stylized proxies: they ask whether a detector separates human from model text under controlled conditions, and rarely specify the institutional setting, the role, the evidence available at decision time, or the consequence of an error. Static corpora also age quickly as models, prompts, and human editing habits change. Transformation-based benchmarks add valuable stress tests, but their labels remain variants of human, machine, source model, attack type, span boundary, or edit magnitude. None of these is the label a chair needs.

\subsection{Pipeline benchmarks}

We propose policy-conditioned pipeline benchmarks. The benchmark object should not be a fixed collection of artifacts but a reproducible recipe for producing artifacts under specified human--AI workflows. A recipe begins from a human seed---a draft, outline, set of notes, manuscript, bug report, or codebase---applies a controlled AI-use pipeline, and assigns a label relative to a stated policy.

Appendix~\ref{app:review} specifies such a recipe set for peer review, with the evidence each workflow leaves behind. Two properties of that set carry the argument. A review that was language-edited, one expanded from the reviewer's own notes, and one written wholly by a model can read alike while the policy separates them, so an artifact-only detector is asked to recover a distinction the artifact does not encode. And one workflow is left undefined by the policy rather than permitted or prohibited by it, so a benchmark that forces such cases to one side reports an accuracy its labels do not support.

A recipe is more durable than a fixed corpus because it can be regenerated: when a new model or paraphraser appears the recipes rerun, and when a venue changes its policy the same pool is relabeled. A complete specification names the seed distribution, the domain, role, and task, the governing policy, the permitted and prohibited pipelines, the tools and prompts, the human writing and verification steps, and the evidence exposed to the detector. That last variable should be varied deliberately across four regimes: $E_0$, the artifact alone; $E_1$, the artifact with a structured disclosure; $E_2$, the artifact with provenance and approved-tool records; and $E_3$, a substantially observed workflow. Uncertainty about the workflow falls across the four, and the inferential burden falls with it. A benchmark that reports performance by regime therefore answers a sharper question than whether workflow-aware detection works: it measures how much each evidence channel buys, and locates the point where inference gives way to verification.

\subsection{Evaluation criteria}

Policy-conditioned benchmarks require more than accuracy or AUROC. AUROC measures ranking across thresholds and does not define a usable procedure. In the review setting a detector may separate fully model-written reviews from hand-written ones on average while still flagging permitted language editing at a rate the venue would never accept.

Evaluation should therefore separate level from power. Permitted workflows define the null and prohibited workflows the alternatives; the level bounds false accusations against compliant reviewers, and power measures detection of specified violations. A detector should not be credited for catching model-written reviews if it also flags permitted editing, and one with valid level but negligible power is safe and useless. Benchmarks should report TPR at fixed FPR per workflow, and should state the base rate they assume, because an operating point on its own does not fix what a flag means.

Suppose a venue receives $20{,}000$ submissions and deploys a detector at TPR $=0.80$ and FPR $=0.01$, a strong operating point on current benchmarks. A flag is an allegation, so the two rates land on two different groups of authors. Take a violation rate of $5\%$. The detector flags $800$ of the $1{,}000$ authors who broke the rule and misses $200$; separately, it flags $190$ of the $19{,}000$ who complied, so four in five flags are justified. Now take a violation rate of $1\%$. The same detector catches $160$ violators and accuses $198$ compliant authors, and most of the people it names did nothing wrong. Loosen the threshold to FPR $=0.05$ and $990$ compliant submissions are flagged against $160$ violations. AUROC is unchanged across the three settings; what changes is the share of accused authors who are innocent, from roughly one in five to roughly six in seven. Reviews inherit the problem, with the added constraint that a flag must be resolved within the review cycle, since discarding a flagged review leaves the paper short of its required count.

Further criteria follow from deployment: calibration of reported probabilities, abstention rate and quality, robustness to paraphrase and deliberate evasion, generalization across models and venues, usefulness to the human adjudicator, and the privacy cost of the evidence required. The last is a constraint rather than a footnote. Workflow evidence is informative precisely because it exposes intellectual process, and a venue that logs reviewer keystrokes to catch delegation has traded one governance failure for another.

Robustness deserves its own arm rather than a line in a list. Any enforced rule invites evasion, and publishing has an attack surface of its own: hidden prompts embedded in manuscripts to steer AI-assisted reviewers \citep{nikkei2025hiddenprompt}, now prohibited at NeurIPS \citep{neurips2026handbook}, and paraphrase attacks on detectors \citep{sadasivan2023,hu2023radar,koike2024outfox}. A benchmark should include recipes in which the actor knows a detector is running and takes low-cost steps to defeat it. Reporting only non-adversarial TPR overstates what deployment will achieve, and the overstatement is largest for the actors a venue most wants to catch.

\section{Governance: What Venues Would Have to Build}\label{sec:governance}

Policy-conditioned detection is only as good as the evidence a venue retains, and that evidence has to come from somewhere. Three things would have to be built: a disclosure format precise enough to condition on, a route by which reviewers reach approved tools without breaching confidentiality, and a process that turns a report into a decision the accused can contest. Building them does not require progress in detection. A venue that has them can enforce a rule with weak inference, while a venue that deploys a strong detector into an uninstrumented pipeline has a score and no procedure.

\paragraph{Disclosure as evidence rather than formality.}
Venue requirements today ask for free text, and only once use crosses a threshold the discloser judges for themselves \citep{neurips2026handbook,iclr2026author}. Categorical schemes exist elsewhere: Amazon KDP separates AI-generated from AI-assisted content and attaches disclosure only to the former \citep{kdp2026}, and the EU AI Act ties obligations to specified content types \citep{euaiact2024}. A declaration naming the task, the tool, the stage, and the person who verified the result changes the question from reconstructing a workflow to checking whether the artifact is consistent with the one the actor described. Watermarks, provenance metadata, and portal edit history enter the same report as further channels, each carrying its own reliability \citep{kirchenbauer2023,dathathri2024synthid,c2pa2026}. None of this survives if a venue leaves open how declarations will be used, since anyone who suspects that declaring invites scrutiny will declare less.

\paragraph{Confidentiality as a constraint on evidence.}
Peer review is confidential, which constrains both the violations that matter and the evidence a venue may gather. Several of the prohibited reviewer workflows in Appendix~\ref{app:review} are violations because a manuscript left the venue's trust boundary, not because the resulting review reads oddly, and the traces that would settle the question are the ones most invasive to collect. Approved-tool routing answers that trade, and it is already in force: NeurIPS 2026 permits reviewers only a venue-sanctioned LLM, and only on papers where LLM use is specifically allowed \citep{neurips2026handbook}. Routing turns an unobservable act into a logged one without surveilling the reviewer's own machine, and it narrows the inference problem, since unexplained model-like text then indicates unsanctioned tooling rather than AI use as such. Strong provenance does not make the framework unnecessary; it changes the kind of evidence the framework runs on. Where a trusted log records a policy-relevant action directly, the question is verification rather than statistical reconstruction. Other parts of the same workflow stay unobserved, including any use of tools outside the venue's infrastructure, and those still require inference. What no venue yet does is treat these logs as evidence inside a stated procedure. A sanctioned-tool record supports a finding only once the venue has also fixed the error rate it will tolerate against compliant reviewers and the route by which a finding can be contested.

\paragraph{Due process and a sanction ladder.}
A report should enter a process rather than settle a decision. Detector errors concentrate on non-native English writers \citep{liang2023}, and detectors fail in ways their users cannot anticipate \citep{weberwulff2023,openai2023classifier}, so a regime that acts on a score alone imposes those errors on identifiable groups. A workable design routes reports through graduated responses: no action, a request for clarification, a structured disclosure request, human integrity review, and sanction, with the evidentiary standard rising at each step and the report never sufficient on its own at the last. Peer review already runs staged and contestable decisions \citep{shah2022peerreview}, and AI-use findings should travel through that machinery rather than around it. The report must also be disclosable to the accused, which requires it to state the hypotheses and calibration set out in Section~\ref{sec:stat_inf}. An inferential report can be answered. A score cannot.

\section{Related Work}\label{sec:related_work}

\textbf{Origin detectors.} One line of work scores text using statistical, likelihood, rank, perturbation, regeneration, or cross-model signals \citep{gehrmann2019gltr,mitchell2023detectgpt,bao2024fastdetectgpt,su2023detectllm,yang2024dnagpt,hans2024spotting}; another trains classifiers on labeled human and machine examples \citep{guo2023hc3,verma2024ghostbuster}; robustness work studies paraphrasing, adversarial rewriting, evasion, and benchmark attacks \citep{sadasivan2023,hu2023radar,koike2024outfox,dugan2024raid}. These produce evidence about origin or model-likeness; the target is detection, and the output is a score.

\textbf{Beyond binary labels.} Editing-sensitive and mixed-authorship methods treat AI involvement as graded or compositional, and watermarking and provenance systems supply evidence about generation history \citep{thai2026editlens,mao2024raidar,wang2024semeval,kirchenbauer2023,dathathri2024synthid,liu2024watermarksurvey}. These are closer to real workflows, and they still require policy interpretation: evidence that a model contributed does not settle whether the contribution was allowed, disclosed, verified, or compatible with the actor's role.

\textbf{AI in peer review.} A growing empirical literature measures model involvement in review rather than in papers. \citet{liang2024monitoring} estimate the prevalence of model-modified review text at AI venues, \citet{latona2024lottery} report score and acceptance effects associated with AI-assisted reviews, \citet{liang2024feedback} evaluate model-generated feedback on manuscripts, and \citet{shah2022peerreview} surveys the computational problems in peer review into which this one falls. Venues themselves now deploy review-side AI \citep{iclr2025feedback} and run experiments with AI authors and reviewers under explicit disclosure \citep{agents4science2025}. This work establishes prevalence and effect; it does not supply a compliance estimand, which is the gap we address. Alongside it, reliability studies report detectors degrading under paraphrase and domain shift, OpenAI's retirement of its own classifier, and bias against non-native English writers \citep{sadasivan2023,openai2023classifier,weberwulff2023,liang2023}. Those findings motivate calibrated evidence with stated validity conditions, abstention, and human review in place of verdicts.

\section{Objections and Limitations}\label{sec:discussion}

\textbf{Origin detection remains valuable.} A natural objection is that many settings still require generic origin detection, and this is correct. Spam and bot detection, platform integrity, synthetic media labeling, and some publishing workflows turn on whether content was machine-generated. Our claim is the narrower one that origin detection constitutes the special case in which the governing policy is ``no AI-generated content,'' rather than the general problem. As AI assistance becomes routine in writing, programming, and research, that special case will describe a diminishing share of the decisions institutions face.

\textbf{The proposal is hard to realize.} Policy-conditioned inference is more demanding than origin classification, and we do not dispute the point. Policies are ambiguous, workflows are only partly observable, actors conceal tool use, and the available evidence is noisy, privacy-sensitive, and open to manipulation. We read these difficulties as an account of why the problem remains unsolved rather than as a defense of the present target. If compliance depends on institutional rules and partly observed workflows, an ``AI detected'' output is aimed at the wrong quantity, and raising its accuracy does not bring it closer to the right one.

\textbf{Data and labels.} The framework requires examples of workflows, not only artifacts: seeds, prompts, drafts, model outputs, revision histories, tool logs, disclosures, and final submissions, labeled by policy status. Collecting them is expensive and raises governance concerns of its own, since workflow evidence exposes intellectual process and confidential material, so evidence acceptable for research and evidence acceptable to collect in deployment are not the same set. Labeling is also partly interpretive: venues will classify the same workflow differently, real workflows fall between editing and drafting, and labels may need to be graded rather than binary.

\textbf{Out of scope.} Decisions taken under this regime feed back into the corpus, since reviews and accepted papers become training data and text shaped by detection pressure will shape later models \citep{shumailov2024collapse}. We do not model that loop. Our worked example is also text-centric. Code, figures, data, and multi-agent research pipelines raise evidence questions we do not address, and the arithmetic assumes one detector applied uniformly, whereas venues will triage.

\section*{Conclusion}

AI-use detection should not be framed as the search for a boundary between human and machine authorship. Venues have written rules that turn on role, task, delegation, and disclosure, and what their enforcement machinery must establish is whether a given workflow complied. Policy-conditioned detection makes that rule an explicit input and treats artifact signals, disclosures, provenance, and workflow traces as evidence of different strengths. Where the workflow stays latent it asks for calibrated inference; where trusted provenance settles the relevant action it reduces to verification. Either way the output has to support a contestable institutional decision at a false positive rate fixed in advance.

The cost of this reframing is that it moves work off the detector and onto the venue. Someone has to write the policy in a form a system can condition on, decide which traces are worth retaining and which are too invasive to collect, fix the tolerated rate of false accusation before any case arrives, and build a path by which a finding can be contested. None of that is a modeling problem, and a better classifier does not supply any of it. But a venue that does it can say what its procedure controls and what it does not, which is the minimum any institution should be able to say before acting on a suspicion. An institution that acts on a detector output owes the accused an account of what was tested and how often that test is wrong. Producing that account is the work this framework describes.

\bibliographystyle{plainnat}
\bibliography{main}

%%%%%%%%%%%%%%%%%%%%%%%%%%%%%%%%%%%%%%%%%%%%%%%%%%%%%%%%%%%%
\newpage
\appendix

\section{A Protocol for Reviewer-Side AI Use}\label{app:protocols}

This appendix instantiates the framework end to end for journal and conference peer review. We state a representative policy, but the partition into permitted, gray-zone, and prohibited workflows will differ across venues, and the protocol should be re-derived from whichever policy is actually in force.

The broader point is procedural. Work that claims to detect or evaluate AI use should, where possible, follow this shape: state the policy, enumerate the workflows the policy permits and prohibits, state the inferential target, and evaluate under evidence conditions that resemble the intended deployment. Such protocols do not remove ambiguity, but they make the resulting claims interpretable and comparable.

\subsection{Policy, inference, and evaluation}\label{app:review}

\subsubsection{Policy}
AI use is permitted when it supports the reviewer's own assessment without replacing confidential judgment, breaching confidentiality, or producing the recommendation. Permitted workflows:

\begin{itemize}
    \item \textbf{Human-only review:} the reviewer reads the manuscript and writes the review without AI. \emph{Evidence:} final review, portal edit history.
    \item \textbf{Model language editing:} the reviewer writes the review first; the model corrects grammar, clarity, tone, or organization. \emph{Evidence:} reviewer draft, edited review, tool log.
    \item \textbf{Model summary support:} the model summarizes non-sensitive parts of the manuscript or the reviewer's own notes, and the reviewer writes the evaluation and recommendation. \emph{Evidence:} reviewer notes, summary prompt, final review.
\end{itemize}

One workflow is treated as gray-zone:

\begin{itemize}
    \item \textbf{Model critique expansion:} the reviewer writes brief notes or bullet-point concerns and the model expands them into a full review. This may preserve the reviewer's judgment, and it may also introduce unsupported criticism, shift emphasis, or create the appearance of delegated evaluation. \emph{Evidence:} bullet notes, expansion prompt, expanded review.
\end{itemize}

Unless the venue resolves it explicitly, this should be a separate class rather than being forced to one side. AI use is prohibited when it delegates evaluative judgment, breaches confidentiality, or introduces claims not grounded in the manuscript:

\begin{itemize}
    \item \textbf{Model-written review:} the model writes the full review from the manuscript. \emph{Evidence:} final review, upload log, absence of any reviewer draft.
    \item \textbf{Model-assigned recommendation:} the model sets the accept/reject recommendation or numerical scores. \emph{Evidence:} scores, submission timing, absence of a reasoning trace.
    \item \textbf{Unsupported criticism:} the model introduces criticisms or factual assertions the manuscript does not support.
    \item \textbf{Unapproved confidential upload:} the reviewer enters confidential manuscript content into an unapproved external tool.
\end{itemize}

\subsubsection{Inference}
In a hypothesis-testing formulation,
\[
H_0: \text{the review was produced through a workflow permitted by the stated review policy,}
\]
\[
H_1: \text{the review was produced through a workflow prohibited by that policy.}
\]
The detector reports whether the available evidence suffices to reject $H_0$, with the uncertainty and evidence conditions under which the claim holds. In a Bayesian formulation it estimates
\[
P(\text{policy violation} \mid \text{review}, \text{manuscript}, \text{review policy}, \text{available workflow evidence}),
\]
and returns a structured report: estimated probability of violation, uncertainty, most likely workflow, evidence used, evidence missing, and a recommendation drawn from no concern, insufficient evidence, human review needed, or likely violation.

\subsubsection{Evaluation}
The benchmark is built from a shared pool of manuscripts or excerpts, each with review criteria, venue policy, and an assigned reviewer role. Multiple workflows are generated from the same manuscript so that evaluation isolates workflow differences rather than manuscript differences.

\begin{enumerate}
    \item \textbf{Human-only review.} Collect manuscripts; ask reviewers to read and write reviews without AI; record scores, recommendation, strengths, weaknesses, and summary.
    \item \textbf{Model language editing.} Start from human-written reviews; prompt the model to correct grammar, clarity, tone, and organization; prohibit new criticisms, score changes, or new claims.
    \item \textbf{Model summary support.} Ask reviewers to write notes after reading; prompt the model to summarize or organize those notes; ask reviewers to write the final review and recommendation; prohibit the model from evaluating the manuscript.
    \item \textbf{Model critique expansion.} Start from reviewer bullet notes; prompt the model to expand them into a full review; allow light reviewer editing. Record whether the expansion added claims absent from the notes.
    \item \textbf{Model-written review.} Start from the manuscript; prompt the model to write a full review; apply light human editing.
    \item \textbf{Model-assigned recommendation.} Start from the manuscript, reviewer notes, or a model-written review; prompt the model to assign scores or a recommendation; optionally have the reviewer copy or justify it.
    \item \textbf{Adversarial arm.} Repeat workflows 5 and 6 with an actor who paraphrases the output, imposes a personal style guide, or splices in hand-written passages to evade detection.
\end{enumerate}

Each item stores the manuscript, the review policy, the workflow type, the policy label, reviewer notes where applicable, prompts and model outputs where applicable, the edited final review, the final scores or recommendation, and the submitted review. Evaluation reports TPR at fixed FPR under permitted workflows, per-workflow breakdowns, calibration of violation probabilities, abstention rate, and performance as a function of the evidence exposed to the detector. The hypothesis to be tested is that workflow-aware inference separates permitted assistance from delegated judgment substantially better than artifact-only detection, and that the gap widens on the adversarial arm.

Detectors carry a documented bias against non-native English writers \citep{liang2023}, and a large share of reviewers at international venues write English as a second language. Evaluation should therefore report FPR under permitted workflows broken out by reviewer language background. A procedure that meets its nominal level in aggregate while exceeding it for one group is not usable for enforcement, whatever its average power.

\section{A Sketch of a Machine-Readable Reviewer Policy}\label{app:policyspec}

The specification below illustrates the minimum content required for a policy that a detection system can condition on and an accused reviewer can contest. Fields are written informally rather than in any particular schema language.

\begin{itemize}
    \item \textbf{Scope:} venue, cycle, role (reviewer), tasks in scope (writing the review text, assigning scores, writing the meta-review).
    \item \textbf{Permitted workflows:} enumerated as in Appendix~\ref{app:review}, each with the tool classes allowed and any confidentiality constraints on inputs.
    \item \textbf{Prohibited workflows:} enumerated, each with the rationale it protects (confidentiality, delegated judgment, unsupported claims).
    \item \textbf{Gray-zone workflows:} enumerated explicitly, with the default handling when observed (for example, disclosure request rather than sanction).
    \item \textbf{Disclosure:} which workflows require disclosure, in what structured form, at what point in the cycle, and how disclosure will be used in adjudication.
    \item \textbf{Evidence:} which traces the venue collects and retains (portal edit history, tool logs, reviewer notes), retention period, and who may access them.
    \item \textbf{Error tolerance:} the maximum acceptable false positive rate against permitted workflows at each rung of the sanction ladder, fixed before the cycle begins.
    \item \textbf{Sanction ladder:} the graduated responses, the evidentiary standard required at each rung, and the statement that a detector report is never sufficient alone for the highest rungs.
    \item \textbf{Appeal:} what is disclosed to the accused party, the response window, and who adjudicates.
\end{itemize}

Writing this down is useful even where no detector is deployed, because it forces a venue to decide in advance which uses it actually intends to prohibit and what rate of error against compliant reviewers it is prepared to accept.

\newpage

\end{document}